\documentclass[letterpaper]{article} 
\usepackage{aaai24}  
\usepackage{times}  
\usepackage{helvet}  
\usepackage{courier}  
\usepackage[hyphens]{url}  
\usepackage{graphicx} 
\usepackage{natbib}  
\usepackage{caption} 
\usepackage{algorithm}
\usepackage{algorithmic}

\usepackage{newfloat}
\usepackage{listings}
\DeclareCaptionStyle{ruled}{labelfont=normalfont,labelsep=colon,strut=off} 
\floatstyle{ruled}
\newfloat{listing}{tb}{lst}{}
\floatname{listing}{Listing}

\usepackage{longtable}
\usepackage{multirow}
\newtheorem{theorem}{Observation}
\usepackage{soul}

\title{Open, Closed, or Small Language Models for Text Classification?}
\author{
    Hao Yu \equalcontrib \textsuperscript{\rm 1,\rm 2},
    Zachary Yang \equalcontrib \textsuperscript{\rm 1,\rm 2},
    Kellin Pelrine \textsuperscript{\rm 1,\rm 2}, 
    Jean Francois Godbout \textsuperscript{\rm 1},
    Reihaneh Rabbany \textsuperscript{\rm 1,\rm 2}
}
\affiliations{
    \textsuperscript{\rm 1}McGill University \\
    \textsuperscript{\rm 2}Mila\\
    \textsuperscript{\rm 3}University of Montreal\\
}

\begin{document}

\maketitle

\begin{abstract} 
Recent advancements in large language models have demonstrated remarkable capabilities across various NLP tasks. But many questions remain, including whether open-source models match closed ones, why these models excel or struggle with certain tasks, and what types of practical procedures can improve performance. We address these questions in the context of classification by evaluating three classes of models using eight datasets across three distinct tasks: named entity recognition, political party prediction, and misinformation detection. While larger LLMs often lead to improved performance, open-source models can rival their closed-source counterparts by fine-tuning. Moreover, supervised smaller models, like RoBERTa, can achieve similar or even greater performance in many datasets compared to generative LLMs. On the other hand, closed models maintain an advantage in hard tasks that demand the most generalizability. This study underscores the importance of model selection based on task requirements.
\end{abstract}

\section{Introduction} 

Recent breakthroughs in large language models have led to significant progress in NLP and text classification. However, many of these models, especially ones with the strongest overall performance like GPT-4, have important limitations such as proprietary restrictions, black box operations, nebulous data sourcing, and high cost and energy consumption. In contrast, smaller, more transparent models might represent a promising alternative, balancing efficacy with sustainability. While comparative evaluations of these models frequently focus on capabilities like understanding, reasoning, and question-answering \citep{Brown2020, llmleaderboard}, they often overlook performance linked to classification tasks. Including these benchmarks is crucial to ensure a more comprehensive evaluation of their strengths and weaknesses across different domains.

So far researchers have shown that closed LLMs like GPT-3.5 and GPT-4 have excellent classification performance in various activities including predicting political party affiliation from social media accounts \citep{tornberg2023chatgpt} and detecting misinformation \citep{pelrine2023reliable}. Unfortunately, evaluations of open-source models for such tasks remain scarce. Do these models measure up to the performance of their closed counterparts, and if so, what specific strategies are needed to achieve this? This study explores eight datasets spanning Name Entity Recognition, political ideology prediction, and misinformation detection. We consider several prompting and tuning techniques to determine the best practices for using LLMs in classification tasks. Our objective is to identify when these models yield strong results and to determine if further refinements are required.

In particular, we compare three sets of representative models from each category: GPT-3.5 and GPT-4 (closed generative LLM), Llama 2 13B and 70B (open generative LLM), and RoBERTa (smaller, non-generative language model). We test the effects of different prompts and zero-shot vs. few-shot vs. fine-tuning setups.

Our main findings are:

\begin{itemize}
    \item Smaller models in a supervised setting can often match or even beat far more costly generative LLMs.
    \item Prompt engineering and other techniques are critical to getting strong results from generative LLMs. With more options such as fine-tuning, we find that open-source models provide an advantage that closed models lack.
    \item The largest, closed models still display a better performance in tasks that are the most challenging and have the strongest demands on generalizability.
\end{itemize}

\section{Related Work} 

The hype around generative LLMs such as GPT-3, ChatGPT, and GPT-4 has grown recently over the results shown in the benchmark datasets, such as Question-Answering, Commonsense Reasoning, and Reading Comprehension. In turn, many researchers have begun to investigate these LLMs' performance on other NLP tasks specific to their domains, contrasting them with established BERT-like models \cite{ye2023comprehensive}. Below, we review their overall capabilities in classification, with a particular focus on Named Entity Recognition (NER), Political Ideology Prediction, and Misinformation Detection, along with a discussion of the limitations of closed-source models.

%

\paragraph{LLMs for classification}

Text classification has significantly evolved over time. Starting from rule-based methods and regexes, NLP later shifted towards classical machine learning methods and then towards deep neural networks. Today, NLP has entered the era of transformer-based models, transitioning from fine-tuning Pre-trained Language Models (PLMs) such as RoBERTa \citep{Kenton2019} to the recent advancements in generative LLMs that require prompt engineering. RoBERTa is an encoder-based model that is pre-trained on the masked language modeling (MLM) task of predicting hidden words in a sentence. In contrast, GPT---Generative Pretrained Transformer---is designed for predicting the next token \citep{Radford2019, Brown2020}. Classification for RoBERTa can be done through feeding the embeddings (last hidden layer) taken from the ``[CLS]'' token and feeding them through a linear layer \citep{Sun2019}.

Generative LLMs show strong comprehension abilities \citep{Liu2023} to human commands, especially after Reinforcement Learning from Human Feedback (RLHF) \citep{Ouyang2022}. Because of this high level of comprehension, these new types of models have led to the creation of a new field called ``prompt engineering'' \citep{AlKhamissi2022} with both manual and automatic prompts to improve task-specific performance \citep{Shin2020}. At the same time, the significance of prompts was exemplified in studies like Chain of Thought \citep{Wei2022} and Tree of Thought \cite{Yao2023}, which can significantly increase performance by suggesting patterns of reasoning for the LLM. Despite the challenge of slight prompt variations causing significant output differences, Instructed GPT, multi-round chat fine-tuning with human feedback aligning \citep{Rafailov2023,Ouyang2022} have attempted to bridge this gap by aligning models with human language patterns. Leveraging the exponential capacity of linearly scaled parameters, LLMs showcased their prowess in complex tasks, including classification. However, many have criticized the tendencies of these models to hallucinate. Guard rails---established for safety and ethical reasons and put in place in the system prompt during RLHF---can be tricked and even jailbroken. Although initial research demonstrated GPT2's classification capabilities with additional classifiers like BERT models, recent focus has shifted towards larger language models with prompt engineering, often leaving classification performance under-reported. 

Another common criticism of LLMs is that their pretraining is resource intensive and almost impossible to execute for most companies and labs. Through Meta's release of Llama \citep{Touvron2023a} and Llama 2 \citep{Touvron2023} to the open-source community, researchers have access to pretrained LLMs to explore how different LLMs perform in various contexts. While smaller, Llama 2 does boast similar capabilities compared to the commercial closed-sourced models of GPT-3.5 and GPT-4, but are still lacking in many areas.
Recently, many researchers have been focused on fine-tuning and inference for these LLMs in low-resource environments and have proposed methods such as fp16, 8bit-Quantized, LoRA \citep{Hu2021} and QLoRA \citep{Dettmers2023}. With these methods, we can dramatically reduce the compute needed to fine-tune open-source pretrained LLMs on various downstream tasks and inject more domain-specific knowledge. In particular, we implement LoRA fine-tuning for Llama 2 on the NER task.


\paragraph{Named Entity Recognition}

Named Entity Recognition (NER) remains a fundamental task in NLP, essential for transforming unstructured text into structured data. This extracted information enhances interpretability in various contexts and feeds into downstream models like Graph Neural Networks (GNNs).

Like other NLP tasks, prevailing NER methods utilize PLMs such as RoBERTa. These models initially extract contextual representations (last hidden outputs of tokens) and employ sub-modules like MLPs \citep{yadav-bethard-2018-survey}, BiLSTMs \citep{Graves2013}, CRF \citep{Souza2019a}, and Global Pointers \citep{Su2022} to aid in entity extraction, and thereby boosting overall performance. Regardless of the method employed, effective NER using these structures depends on well-annotated datasets for effective fine-tuning to achieve optimal performance.


Recent studies by \citet{Li2023,Wang2023a} have shown promising results when incorporating GPT-3.5 for NER. These models were prompted to generate specific entities in desired formats, and this was followed by straightforward post-processing. Furthermore, they demonstrated that robust zero-shot and few-shot capabilities significantly improved performance, even in contexts with limited resources. Based on these findings, we propose adopting a similar approach in several experiments presented in the second part of the paper. 

\paragraph{Political Ideology Prediction}

Political ideology prediction is the first of many steps computational scientists conduct to analyze partisan discourse or polarization. This task is typically framed as predicting the party or ideology of social media users. The literature references a wide variety of features ranging from textual content \citep{conover2011predicting, rodriguez2022urjc, mou2021align, fagni2022fine}, various types of network information \citep{Barber2015TweetingFL, colleoni2014echo, pennacchiotti2011machine, gu2016ideology, xiao2020timme, havey2020partisan, wojcieszak2022most, jiang2021social} to other features like ideology of well-known media outlets from which users share stories \citep{rheault2021efficient, luceri2019red, stefanov2020predicting, badawy2018analyzing}. The Authors' forthcoming ICWSM paper (anonymized) provides a comparative survey and empirical analysis of various domain-specialized and non-generative approaches, showing that RoBERTa achieves strong performance equal or superior to specialized models in this particular task.

Previously, human labels have been the gold standard unless self-declared labels are available (e.g., politicians or users responding to a survey). But \citet{tornberg2023chatgpt} showed that GPT-4 performed better than human annotators, even experts, in determining the party affiliation of politicians from their messages on social media. While a strong result, this leaves open questions that motivate our work here. First, prior research shows that politician behavior can be different and sometimes easier to predict than the general public \citep{cohen2013classifying}. Second, \citet{tornberg2023chatgpt} focused on the United States, which has a two-party system. Identifying party ideology in a multiparty system might be more challenging. Finally, this task is generally used as a foundation for downstream research, and cost may be an issue, as well as other concerns like models changing over time. Here, we address these questions by testing approaches to classifying the general public in both the US (two-party) and Canada (multi-party).

\paragraph{Misinformation Detection}
Misinformation is a critical societal challenge to which a great deal of research has been devoted \citep{shu2017fake,kumar2021battling,shahid2022detecting}. One of the main tools aimed at countering the spread of fake news is algorithmic detection, usually framed as a classification problem (e.g., labeling information as ``True'' or ``False'') \citep{shu2017fake}. While there are many approaches based on network information or user profiling, textual content is central, and often the only way a prediction could be made with certainty given that content is what actually determines veracity.

Older approaches such as SVM, CNN, LSTM, etc. were once prevalent \citep{shu2017fake}, but transformer-based language models such as BERT have generally been shown to provide superior performance in detecting misinformation \citep{pelrine2021surprising, kaliyar2021fakebert}. More recently, GPT-4 gave even stronger performance and other benefits like better generalization and uncertainty quantification \citep{pelrine2023reliable}. However, with massive amounts of potential misinformation created every day, scalability remains a key challenge, and GPT-4 is expensive and strictly rate-limited. To our knowledge, it has not yet been determined if recent scalable open-source generative LLMs---like Llama 2---are effective in this domain. Testing these newer models is also one of our objective. 

\paragraph{Limitations of Close-Sourced Models}

Closed-source models such as GPT-3.5 and GPT-4 boast impressive performance across various NLP tasks; however, they are accompanied by several limitations. Typically, these models are accessed through APIs, relieving users of computing infrastructure concerns. Although they are user-friendly, cloud based AI services lack control over training data and model versioning. The undisclosed nature of the training corpus makes it challenging to determine whether a model's success on benchmark datasets is due to effective generalization or potential data leakage. Moreover, reproducing research conducted on closed-source models proves difficult due to the high cost associated with running experiments via APIs (considering GPT-3.5 \& GPT-4 costs) and unanticipated model updates \citep{pozzobon2023challenges}, which can lead to fluctuating performance \citep{chen2023chatgpts}. In addition, many of these closed-source models incorporate interactions with APIs into their subsequent model's training dataset, raising ethical and privacy concerns. Finally, the significant energy consumption required to train and run these LLMs has a substantial environmental impact, making their use a concern for sustainable practices.

\section{Methodology} 

In the following sections, we describe the common models evaluated and the three main tasks. For each task, we describe the dataset, their setup and evaluation metrics used. We provide further technical information such as the code and exact prompts in the supplementary material.

\subsection{Models}
For our experiments, we compare the performance between two popular generative LLMs against the state of the art methods. In particular, we compare the open-sourced model Llama 2 Chat  \citep{touvron2023llama} against the closed source models of GPT-3.5 \citep{brown2020language} and GPT-4 \citep{openai2023gpt4}. 
For the state of the art methods, we fine-tune RoBERTa \citep{liu2019roberta} to perform the various classification tasks \cite{Wang2020}. More specifically, we use ``Llama-2-13b-chat-hf'' and ``Llama-2-70b-chat-hf'' hosted on HuggingFace. For GPT-3.5 and GPT-4, we use OpenAI's API with the model parameter specified as ``gpt-3.5-turbo-0613'' and ``gpt-4-0613'' respectively. These generative LLMs are optimized for dialogue use, allowing us to use the same prompts. Table \ref{tab:model_comparison} shows the different model size, type and training. 

\begin{table}[ht!]
\centering
\begin{tabular}{l|lll}
Model &  Size & Type & Training \\ \hline
Llama 2 & 13B, 70B & Decoder & Unsupervised \\
GPT-3.5 & 175B & Decoder & Unsupervised \\
GPT-4 & 220B - 1.76T & Decoder & Unsupervised \\
\hline
RoBERTa & 123M, 354M & Encoder & Supervised 
\end{tabular}
\caption{Model Comparison. GPT-4 model parameter size is estimated: \url{https://the-decoder.com/gpt-4-is-1-76-trillion-parameters-in-size-and-relies-on-30-year-old-technology/}}
\label{tab:model_comparison}
\end{table}

\paragraph{Llama 2 Inference Setup}
We use vLLM \citet{WoosukKwon2023}, a fast and highly efficient library for LLM inference across 2 $\times$ A100 80GB GPUs. 
For Llama 2 (70B), we set the requests per minute at 500 and tokens per minute at 120,000.

\paragraph{Zero-shot \& Few-Shot}
For our generative LLMs, we evaluate each task on the zero-shot and few-shot setting. In the zero-shot setting, only instructions about the task are provided. In the few-shot setting, two examples per class from the training set are provided as past conversations, where the ``user'' role has the text to classify and the ``assistant'' role provides the expected answer.

\subsection{Classification Tasks}
We consider three text classification tasks: NER, classifying political party and detecting misinformation. For comparison, we describe the number of classes and the size of the training and test sets in Table \ref{tab:dataset_metadata}. 

\begin{table}[ht!]
\small
\centering
\begin{tabular}{ll|crr}
Task & Dataset & Classes & Train & Test \\ 
\hline
\multirow{3}{*}{NER} & CoNLL 2003 & 4 & 35,350 & 3,453 \\ 
                     & WNUT 2017 & 6 & 3,394  & 1,287 \\ 
                     & WikiNER-EN & 4 & 115,473  & 14,435 \\ 
\hline
\multirow{3}{*}{Ideology} & 2020 Election & 2 & 1,141  & 356 \\ 
                    & COVID-19 & 5 & 2,013  & 629 \\ 
                    & 2021 Election & 5 & 2,060  & 643 \\ 
\hline
\multirow{2}{*}{Misinfo} & LIAR & 2 & 10,269  & 1,283 \\ 
                         & CT-FAN-22 & 3 & 900  & 612 
\end{tabular}
\caption{Dataset metadata.}
\label{tab:dataset_metadata}
\end{table}

\paragraph{NER \label{NER}} 
On the NER task, we evaluate and report the F1-score on the subset of the official test split of three common benchmarks:  CoNLL 2003 \citep{tjong-kim-sang-de-meulder-2003-introduction}, WNUT 2017 \citep{derczynski-etal-2017-results}, and WikiNER-EN \citep{nothman2012:artint:wikiner}. 
For the generative LLMs, we test two styles of prompts: ``Serial'' and ``JSON'' in a zero-shot and few-shot setting. Technical details on the prompts and formatting is provided in the supplementary materials. 


We also perform LoRA fine-tuning \citep{hiyouga2023} of Llama2 (70B) to detect the ``PERSON'' and ``LOCATION'' entities. For this fine-tuning, we combine the training set of four common NER datasets: CoNLLpp \citep{Wang2019}, WNUT-2017, WikiNER-EN, and OntoNotes5.0 \citep{Pradhan2013}. Table \ref{table:ner_datasets} reports the number of samples in the training and test set. We trained 1 epoch with $5 \times 10^{-5}$ learning rate on 4 $\times$ A100 80GB. For this fine-tuning, it took 17.5 hours for the loss to drop to around $0.02 \pm 0.01$.

\begin{table}[ht!]
\centering
\begin{tabular}{l|lll}
Dataset & Train & Test & Entity \\
\hline
CoNLL 2003  & 14041  & \textbf{3453}  & B/I-PER       \\ 
WNUT 2017   & {\underline {3394}}  & \textbf{1287}  & B/I-person \\ 
WikiNER-EN & {\underline{115473}} & \textbf{14435} & B/I-PER      \\ 
OntoNotes5 & {\underline{12195}}  & 1573           & B/I-PERSON \\
CoNLLpp    & {\underline{14041}}  & 3453           & B/I-PER   
\end{tabular}                          
\caption{Number of samples in each split of NER datasets combined to perform LoRA fine-tuning of Llama 2 (70B). Train sets are underlined and test splits are in bold.}
\label{table:ner_datasets}
\end{table}

\paragraph{Political Ideology Prediction}
For the political ideology prediction task, we examine three datasets collected using Twitter's API: ``2020 (US) Election'', ``(Canada) COVID-19'', and ``2021 (Canadian) Election''. These datasets represent 1\% of real-time tweets collected with their respective keywords (provided in Supplementary). We evaluate two tasks: ``Explicit'' and ``Implicit'' political ideology prediction. The``explicit'' task is to identify the user's political ideology based on their explicit profiles, which are those that contain keywords related to their ideology, i.e. a profile containing ``Joe Biden'' would imply support for or against the US democratic party. Two political scientists were recruited to manually annotate the sampled profiles. Table \ref{tab:political_party_dataset} reports the time frame, number of users, number of tweets, and the inter-annotator score. The ``implicit'' task is to classify users' political ideology solely based on their tweets without their profile information. 

\begin{table}[ht!]
\small
\centering
\begin{tabular}{l|lll}
 & 2020 Election & COVID-19 & 2021 Election \\ 
\hline
Start & 2020-10-09 & 2020-10-09 & 2021-08-01 \\
End & 2021-01-04 & 2021-01-04 & 2021-10-22 \\
Total Users & 23,758,112 & 4,765,115 & 775,607 \\
Total Tweets & 387,090,097 & 231,841,790 & 11,361,581 \\
Labeled Users & 1,782 & 3,145 & 3,217 \\
Cohen Kappa & 0.76 & 0.74 & 0.61 \\
\end{tabular}
\caption{Twitter political datasets.}
\label{tab:political_party_dataset}
\end{table}

For both tasks, we evaluate the same five random seeds on 20\% of the labeled users and report the weighted average F1-score. For the generative LLMs, we test in a zero-shot and few-shot setting and prompt for the exact class labels. For the ``implicit'' task, we include as many tweets as possible within the max token size. For the smaller supervised models, we do the following: On the ``explicit'' task, we fine-tune RoBERTa-large on the training set. On the ``implicit'' task, we first pre-train RoBERTa-base on all tweets from the respective dataset for 1 epoch. We then produce a 768 embedding of each tweet with this pre-trained RoBERTa model and create a user embedding by the mean aggregation of each user's tweet embedding. We then train a two-layer fully connected MLP to predict the user's political ideology. More details are provided in the supplementary materials. 

\paragraph{Misinfo Detection}

For the misinformation detection task, we compare the performance on two datasets: ``LIAR'' \citep{wang-2017-liar} and ``CT-FAN-22'' \citep{kohler2022overview}. For the generative LLMs, we prompt the models to return a truthfulness score between 0-100, where 0 is a blatant lie. In the zero-shot setting, we split the range evenly amongst the classes. In addition, to match prior evaluation on these datasets, we use accuracy for LIAR and macro-F1 for CT-FAN-22. For the smaller supervised models, we fine-tune a RoBERTa-large model, matching the overall best-performing model shown in \cite{pelrine2021surprising}. 

If the model doesn't correctly generate a score, we generate a uniform random one to make sure all results are comparable on the same data. Similarly, for the category "other" in CT-FAN-22 that is ill-defined and cannot be obtained from a score nor any other prompting we are aware of, we directly mark the examples as incorrect predictions, providing the most stringent evaluation of generative LLM performance here. Following the most common practices on the two datasets, we evaluate accuracy on LIAR (which has near-balanced classes) and macro F1 on CT-FAN-22.


\section{Results}

In this section, we present the outcomes of our experiments. We first analyze the aggregated results, comparing the performance of the two generative LLMs and contrasting them with RoBERTa. Next, we demonstrate that with fine-tuning, open-source LLMs can also outperform GPT-3.5. Subsequently, we explore specific scenarios to further discuss when smaller discriminative models are preferred over generative LLMs and vice versa in classification tasks. Our section concludes with a cost analysis.

\subsection{Llama 2 vs. GPT-3.5 vs. RoBERTa}

Table \ref{tab:ALL_NLP_INFO} shows the best test score attained from prompting each generative LLMs in a zero-shot and few-shot setting. The final column provides the test score attained through fine-tuning a RoBERTa model. 
To reduce our total cost incurred by GPT-4, we only run on the most challenging dataset for the NER classification task and implicit political party prediction task. The following paragraphs detail our observed findings.

\begin{table*}[ht!]
\centering
\begin{tabular}{ll|llll|l}
Task & Dataset & Llama 2 (13B) & Llama 2 (70 B) & GPT-3.5 & GPT-4 & RoBERTa \\ \hline
\multirow{3}{*}{NER} & CoNLL 2003 & 57.8 ± 11.5 & \underline{82.5 ± 5.6} & 79.8 ± 6.2 & -- & \textbf{94.3 ± 3.5} \\
                     & WNUT 2017 & 35.4 ± 4.7 & 55.3 ± 4.7 & 54.6 ± 3.0 & \textbf{65.1 ± 3.0} & \underline{59.6 ± 3.3} \\
                     & WikiNER-EN & 51.3 ± 8.8 & 76.1 ± 3.6 & \underline{77.4 ± 0.6} & -- & \textbf{96.2 ± 0.1} \\
\hline
\multirow{3}{*}{Explicit Ideology } & 2020 Election & 95.5 ± 1.1 & 96.3 ± 0.5 & 97.0 ± 0.8 & \textbf{97.6 ± 0.5} & \underline{97.3 ± 0.6} \\
                                    & COVID-19 & 90.2 ± 0.9 & 92.5 ± 1.3 & \underline{94.7 ± 0.8} & \textbf{95.1 ± 0.6} & 91.2 ± 0.2 \\
                                    & 2021 Election  & 82.1 ± 1.6 & 85.2 ± 1.0 & 87.7 ± 1.3 & \underline{89.4 ± 1.2} & \textbf{95.2 ± 0.7} \\
\hline
\multirow{3}{*}{Implicit Ideology } & 2020 Election & 71.9 ± 1.9 & 77.2 ± 1.0 & \underline{92.9 ± 0.5} & -- & \textbf{93.0 ± 0.2} \\
                                    & COVID-19 & 44.6 ± 1.6 & 53.9 ± 1.5 & 65.9 ± 2.0 & \underline{68.6 ± 1.9} & \textbf{70.0 ± 2.7} \\
                                    & 2021 Election & 48.8 ± 3.5 & 55.7 ± 3.3 & \underline{75.4 ± 1.6} & -- & \textbf{82.3 ± 1.1} \\ 
\hline
\multirow{3}{*}{Misinfo} & LIAR & 50.0 ± 1.3 & 49.1 ± 2.5 & \textbf{68.5 ± 3.0 } & \underline{66.3 ± 2.1} & 61.5 ± 2.1\\
                         & CT-FAN-22 & 21.2 ± 3.2 & 25.4 ± 2.1 & \textbf{43.7 ± 1.9} & \underline{42.0 ± 2.6} & 21.6 ± 2.0
\end{tabular}
\caption{Performance of Generative LLMs for NER, Explicit and Implicit Political Ideology Prediction, and Misinformation Detection. For LLMs, we report the best score achieved across zero-shot and few-shot settings}
\label{tab:ALL_NLP_INFO}
\end{table*}



\begin{theorem}
Smaller models in a supervised setting often reach similar performance or outright outperform generative LLMs.
\label{obs:small_supervised_model_wins}
\end{theorem}

In Table \ref{tab:ALL_NLP_INFO}, we find that RoBERTa often outperforms both versions of Llama 2, and GPT-3.5. There is only one task/dataset where it does worse by a large margin (CT-FAN-22 -- still better than Llama, but worse than GPT-3.5). In a number of cases it even beats GPT-4. Considering the substantial advantages of RoBERTa in cost, speed, transparency, and more, it remains well worth considering and the superior choice in many applications.

\begin{theorem}
Prompts that work well in zero-shot setting do not necessarily work well for few-shot setting.
\label{obs:prompt_engineering}
\end{theorem}

Observation \ref{obs:prompt_engineering} can be observed from Table \ref{tab:ner_result} when comparing between the two prompting styles: ``Serial'' and ``JSON''. Our findings underscore the significance of prompt engineering. Specifically, we demonstrate that, when applying the same prompt to different models (Llama 2 and GPT), it is not guaranteed to yield the same results. Although both models share a ``chat'' structure, comprising ``system'', ``user'', and ``assistant'' roles, our observations indicate that LLama 2 performs better with the ``Serial'' style in the zero-shot scenario. Conversely, GPT-3.5's performance remains consistent whether prompted in ``Serial'' or in ``JSON'' mode. Notably, in the few-shot setting, we discovered that only the ``JSON'' prompt offers advantages for both models, whereas the ``Serial'' prompt leads to a decline in performance. We hypothesize that providing examples in common data structures could enhance model comprehension, potentially benefiting from prior training with reinforcement learning from human feedback (RLHF) to grasp such structures more effectively.

\begin{table*}[th!]
\begin{tabular}{lc|lll|lll}
\multirow{2}{*}{Model} & \multirow{2}{*}{Prompt} & \multicolumn{3}{c|}{Zero-Shot} & \multicolumn{3}{c}{Few-Shot}  \\
 &  & CoNLL2003 & WNUT2017 & WikiNER-EN & CoNLL2003 & WNUT2017 & WikiNER-EN \\ \hline
Llama 2 (13B) & \multirow{4}{*}{Serial} & 52.4 ± 7.4 & 32.4 ± 5.5 & 40.6 ± 2.4 & 24.0 ± 3.5 & 19.5 ± 3.7 & 30.6 ± 5.6 \\
Llama 2 (70 B) &  & 63.4 ± 7.2 & 43.4 ± 3.8 & 60.3 ± 7.3 & 9.1 ± 7.1 & 23.0 ± 4.4 & 16.3 ± 3.3 \\
GPT 3.5 &  & 79.3 ± 7.2 & 47.2 ± 7.9 & 77.4 ± 0.6 & 55.2 ± 11.6 & 49.9 ± 4.7 & 54.5 ± 5.3 \\
LoRA Llama 2 &  & 80.5 ± 5.3 & 58.7 ± 4.1 & 72.2 ± 5.4 & 0.0 ± 0.0 & 0.0 ± 0.0 & 0.0 ± 0.0 \\ \hline
Llama 2 (13B) & \multirow{4}{*}{JSON} & 42.0 ± 8.6 & 28.7 ± 4.8 & 46.7 ± 8.2 & 57.8 ± 11.5 & 35.4 ± 4.7 & 51.3 ± 8.8 \\
Llama 2 (70 B) &  & 61.3 ± 11.3 & 46.4 ± 4.2 & 59.9 ± 8.1 & 82.5 ± 5.6 & 55.3 ± 4.7 & 76.1 ± 3.6 \\
GPT-3.5 &  & 75.6 ± 7.9 & 54.6 ± 3.0 & 70.4 ± 5.3 & 79.8 ± 6.2 & 50.9 ± 6.1 & 74.9 ± 4.0 \\
LoRA Llama 2 &  & \underline{88.1 ± 3.7} & \textbf{60.9 ± 7.1} & \underline{87.7 ± 1.5} & \underline {88.1 ± 3.6} & \textbf{60.5 ± 6.4} & \underline {87.5 ± 1.5} \\ \hline
RoBERTa & -- & \textbf{94.3 ± 3.5} & \underline {59.6 ± 3.3} & \textbf{95.7 ± 0.1} & \textbf{94.3 ± 3.5} & \underline {59.6 ± 3.3} & \textbf{95.7 ± 0.1}
\end{tabular}
\caption{NER: Serial vs. JSON prompting under Zero-Shot and Few-Shot setting.}
\label{tab:ner_result}
\end{table*}

\begin{theorem}
With access to open-source models, we can further train the model to beat closed-source models.
\end{theorem}

We analyze data from Table \ref{tab:ner_result}. Due to a noticeable contrast between Llama 2 and GPT-3.5, we aimed to enhance Llama 2's performance through some fine-tuning. To achieve this, we subjected the Llama 2 (70B) model to fine-tuning using LoRA. While this fine-tuning enabled us to outperform GPT-3.5, it did not bridge the gap with the smaller model like RoBERTa.

\subsection{Generalizability}

\begin{theorem}
Supervised smaller models can leverage non-semantic training data patterns, but become specialized and lose generalizability. If one wants to take advantage of such patterns, they can be effective, while if one needs broader generalizability then larger generative models may be preferable. 
\end{theorem}

This observation is based on the following analyses of errors and dataset difficulty.

\paragraph{Canadian Implicit Ideology Prediction}

In Table~\ref{tab:ALL_NLP_INFO}, we observe that generative models perform relatively poorly on implicit ideology prediction for Canadian data, especially the 2021 Election dataset. Where does RoBERTa's advantage come from? 

To better understand this, we would like to separate understanding of text in general and Canadian politics specifically, from other patterns that might be learned from the training data (for example, transient patterns in discussion topics or hashtag usage among different groups on Twitter, such as one group using \#ElectionCanada and another \#Election2021). The former two might be improved with better prompts or providing the generative model with overall domain knowledge, while the latter would require a different approach to leverage less semantic patterns that could be found in the training data.

To analyze this, we examine cases from the 2021 Election dataset where GPT-3.5 and RoBERTa predictions disagree, where one was incorrect while the other was correct, and vice versa. We randomly sampled 20 users from each class. In the case where GPT-3.5 was incorrect and RoBERTa was correct, all five classes had users. For the other case, there were no users from the LPC class.

We provided the users' tweets to a political scientist, who first assessed whether a political affiliation could be determined, and then identified the specific party to which the user belonged. Between the profile and tweets, the political scientist had an F1-score of 49.3\%, labeling 76 out of the 180 users the same political party as identified from their profile. For 54 users, the political scientist could not figure out which party the user belonged to. GPT-3.5 could not provide a political party for 16 users.

We examine the relationship further by showing the confusion matrices between the labels from the political scientist based on tweets solely against the answers from RoBERTa and the answers from GPT in Table \ref{tab:implicit_ideology_prediction_confusion_matrix}. Notably, the predictions from GPT-3.5 have a much closer alignment with those of the political scientist who relied exclusively on the user tweets for classification (Cohen Kappa 0.068 for RoBERTa vs. 0.332 for GPT). Consequently, these results suggest that RoBERTa's advantage in this data comes from non-semantic patterns that it is able to find in the training data.

\begin{table*}[ht!]
\begin{tabular}{ll|llllll|llllll}
 &  & \multicolumn{6}{c|}{RoBERTa} &  \multicolumn{6}{c}{GPT-3.5} \\
 &  & None & CPC & PPC & GPC & LPC & NDP &  None & CPC & PPC & GPC & LPC & NDP \\ \hline
\multirow{6}{*}{Human Labels} & None & 0 & 14 & 4 & 3 & 16 & 17 &   7 & 7 & 6 & 27 & 1 & 6 \\
 & CPC & 0 & 9 & 3 & 1 & 9 & 5 &   2 & 17 & 2 & 4 & 0 & 2 \\
 & PPC & 0 & 17 & 17 & 0 & 0 & 2 &   5 & 14 & 17 & 0 & 0 & 0 \\
 & GPC & 0 & 0 & 0 & 0 & 15 & 10 &   0 & 3 & 0 & 12 & 5 & 0 \\
 & LPC & 0 & 1 & 0 & 17 & 5 & 5 &   1 & 2 & 0 & 10 & 10 & 5 \\
 & NDP & 0 & 6 & 0 & 5 & 4 & 5 &   1 & 0 & 0 & 5 & 0 & 14
\end{tabular}
\caption{Confusion Matrices of RoBERTa and GPT-3.5 against the Human Labels of the Tweets.}
\label{tab:implicit_ideology_prediction_confusion_matrix}
\end{table*}

\paragraph{Misinformation Task Difficulty}

RoBERTa performs fairly well when fine-tuned on LIAR---behind zero-shot GPT but not by too much. However, it gives terrible performance on CT-FAN-22. We hypothesize that this is due to increased variation in CT-FAN, which is sourced from 15 fact-checking websites \citep{shahi2021overview, kohler2022overview}, compared with LIAR which is only sourced from one. The increased diversity increases the difficulty of generalization. In fact, fine-tuning on LIAR and then testing on CT-FAN actually gives slightly better performance than fine-tuning on CT-FAN itself (26.8 vs. 23.0 F1), indicating RoBERTa is unable to learn from the training data in CT-FAN.

Generalizability is especially important for this task, as real-world misinformation is very diverse, evolves quickly, and is much harder to label than NER or party prediction examples. Consequently, GPT has a significant advantage here, though RoBERTa might be considered in very constrained, static settings (for example, detecting a specific, known misinformation narrative).

\subsection{Model Cost Analysis}

As LLMs are trained with more and more parameters, many have raised concerns about the substantial computational demands and the accompanying environmental impact, particularly in terms of carbon footprint, energy consumption, and the cost-effectiveness of their outputs. We report the cost incurred for the NER task in Table \ref{table:energy}.

\begin{table*}[ht!]
\centering
\begin{tabular}{l|lllll}

\multicolumn{1}{l|}{\textbf{Training}} & Speed (sample/s) & Training Time (s) & GPU Power (W) & Energy (kWh) & Cost (USD) \\ \hline
RoBERTa & 235 & 3,430 & 966 & 0.921 & 5.87 \\
Llama 2 (70B) & 0.434 & 62,900 & 1430 & 25.0 & 225 \\
\hline
\\
\multicolumn{1}{l|}{\textbf{Inference}} & Speed (sample/s) & Inference Time (s) & GPU Power (W) & Energy (kWh) & Cost (USD) \\  \hline

\multicolumn{1}{l|}{Llama 2 (13B)} & 15.0 & 66.7 &  & 0.0132 & 0.0916 \\
\multicolumn{1}{l|}{Llama 2 (70B)} & 8.33 & 120 &  & 0.0238 & 0.165 \\
\multicolumn{1}{l|}{RoBERTa} & 497 & 2.01 & \multirow{-3}{*}{714.4} & 0.000399 & \textbf{0.00276} \\
\hline
\multicolumn{1}{l|}{} & Av. Prompt Tokens & Av. Comp. Tokens & Prompt Tokens & Completion Token & Cost (USD) \\ \hline
\multicolumn{1}{l|}{GPT-3.5} &  &  &  &  & 0.0441 \\
\multicolumn{1}{l|}{GPT-4 } & \multirow{-2}{*}{405} & \multirow{-2}{*}{22.0} & \multirow{-2}{*}{40500} & \multirow{-2}{*}{2200} & 13.5 \\ 

\end{tabular}
\caption{Summary of energy consumption and cost for training and inferring 1,000 examples across different models.}
\label{table:energy}
\end{table*}

We use Weights \& Biases \cite{Biewald2020} during training and inference. This library logs many experimental results as well as providing advanced monitoring and measurement tools such as recording GPU power utilization and run time. We first calculate the total cost and energy from training the models. Since we fine-tune RoBERTa, we sum the training time of three datasets: CoNLL 2003, WNUT 2017, and WikiNER-EN. For the LoRA fine-tuning of Llama 2, we calculate the training time required for the 1 epoch.  For inference, we calculate the inference time, energy consumption and monetary cost for inferring 1,000 samples. Our prices reflect the local electrical price (\$4.82 USD / kWh) and the price of cloud hardware suppliers (\$ 1.5 USD / A100 80GB / h).


\begin{theorem}
Open-source models are beneficial. Supervised smaller models are the best for the environment.
\label{obs:open_source_is_good}
\end{theorem}

As observed in Table  \ref{table:energy}, RoBERTa uses the least amount of energy for training and inference, as it is the smallest model. Comparing with the results from Table \ref{tab:ALL_NLP_INFO}, RoBERTa can achieve similar performance or even surpass the generative LLMs. As such, when the classification task is simple and the patterns are well-defined, fine-tuning a RoBERTa model would be the best choice. For inference, Llama 2 (13B \& 70B) demonstrate high throughput, but still consumes quite a bit of energy comparatively. Considering the performance, the costs are not favorable.

Among closed models, GPT-3.5 performs well and GPT-4 even better. However, the difference in the cost between the two is typically too high compared to the performance gain. Thus, sticking to GPT-3.5 would be recommended for most real-world applications, if RoBERTa is insufficient. 


\section{Conclusion}

In this research paper, we investigated the performance of various language models in different NLP classification tasks, including NER extraction, political ideology prediction, and misinformation detection. Our analysis involved comparing open-source models such as Llama 2 and closed-source models like GPT-3.5 and GPT-4, as well as supervised models like RoBERTa.

Our findings reveal several key observations. Smaller supervised models, such as RoBERTa, often achieve similar or superior performance compared to generative language models, while offering considerable advantages in terms of cost, speed, and transparency. We also observe the importance of prompt engineering, where the choice of prompts significantly impacts a model's performance. Prompting styles that work well in zero-shot settings do not necessarily yield the same results in few-shot settings, highlighting the complexity of prompt design. Furthermore, by utilizing fine-tuning techniques, we demonstrated that open-source models like Llama 2 could still outperform closed-source models like GPT-3.5, emphasizing the value of collaborative open-source initiatives. It is noteworthy that while supervised models like RoBERTa excelled in tasks where patterns were well-defined, generative models like GPT-3.5 exhibited greater performance in tasks where generalization and transferability were critical.

Our research underscores the significance of selecting the appropriate model based on the task's characteristics, the availability of resources, and the need for generalization. Additionally, our investigation helps better understand the limitations of closed-source models and highlights the potential of open-source models. In return, these insights and findings can foster reproducibility and collaborative research in the NLP domain. As the field of natural language processing continues to evolve, the insights gained from this research can inform the selection and utilization of models to suit the specific requirements of various NLP classification tasks.

\bibliography{bibliography}

\newpage

\section{Supplementary Material: NER}

We separately run the model with a simple post-process for the evaluation dataset to get the entity.
For the fine-tuning step, we combined these datasets (WNUT2017, WikiNER-EN, OntoNotes5, CoNLLpp) and restructured them to fit the dialogue format required by the Chat-optimized model.

Here is one sample of the combined dataset:\\ \emph{
[\{
    \\\indent\textbf{``instruction"}: Extract human names and locations or addresses in json format, like \{``name'': [name 1, name 2, ...], ``location'': [location 1, location 2, ...]\}. Your words should extract from the given text, do not add/modify any other words. Keep your answers as short as possible, remember do not include phone number. For every name, there should be less than 3 words. Only output the json string without any other content.",
    \\\indent\textbf{``input"}: ``Peter Blackburn",
    \\\indent\textbf{``output"}: ``\{``name": [``Peter", ``Blackburn"], ``location": []\}''\\
 \},...]
}
\subsection{Fine-tuned Results}

\begin{figure}[ht]
   \includegraphics[width=1\linewidth, trim={0 0 0cm 1.4cm}, clip]{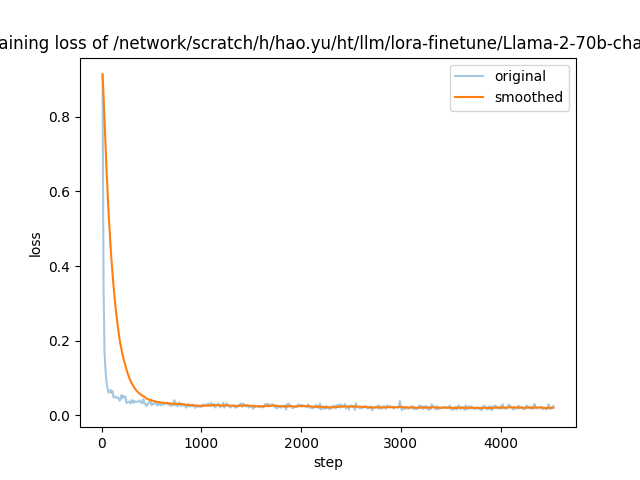}
   \caption{The training loss curve for supervised finetuning with Llama2 70B Chat on the combined dataset.}
   \label{fig:70b-loss}
\vspace{-0.5cm}
\end{figure}

Figure \ref{fig:70b-loss} presents the training loss of the Llama2 Chat (70B)  supervised finetuning (SFT) on our combined datasets. The curve fits the normal training curve; the loss gradually approaches a stable value of around 0.02. The difference between non-finetuned and finetuned in Table 6 in the main paper indicates that the training process works well and Llama 2 learns to do NER. 
 
\subsection{Prompts}

The following prompts are used for all models in our experiments. The underlined parts are the only difference between Serial and JSON prompt styles. We replace the \{\{text\}\} with the respective sentence in the datasets for inference.

\textbf{System Prompt}: \emph{You are a helpful and efficient system only return user desired minimum output. You can't add any other content, words and explanation.}

\textbf{Serial Prompt}: \emph{As an unbiased labeler, please extract people names, locations or addresses. Need follow these rules: Entities should be in the given text. Do not add or modify any words. Keep the entity as short as possible. Do not include any phone numbers. A person's name should be less than 3 words. Separate multiple entities by '$|$' Only generate the output without any other words. \\
Text: \{\{text\}\} \\
Output if entities detected: \ul {Names: name1$|$name2\textbackslash n Locations: location1$|$location2\textbackslash n}\\
Output if not entities detected: \ul {Names: None\textbackslash n Locations: None\textbackslash n}}

\textbf{JSON Prompt}: \emph{As an unbiased labeler, please extract people names, locations or addresses. Need follow these rules: Entities should be in the given text. Do not add or modify any words. Keep the entity as short as possible. Do not include any phone numbers. A person's name should be less than 3 words. Separate multiple entities by '|' Only generate the output without any other words. \\
Text: \{\{text\}\} \\
Output if entities detected: \ul{\{``name": [name 1, name 2, ...], ``location": [location 1, location 2, ...]\}} \\
Output if not entities detected: \ul{\{``name": [], "location": []\}}}

For few-shot, we switch the start of the prompt to be:

\textbf{Serial Few-Shot Examples}:\\
    \emph{ \textbf{user}: The government of Chad has closed N'Djamena University after two days of protests over grant arrears in which Education Minister Nagoum Yamassoum was held hostage for four hours, state radio said on Friday.}\\
    \emph{ \textbf{assistant}: \ul{Name: Nagoum$|$Yamassoum\textbackslash n Locations: Chad\textbackslash n}}\\
    \emph{ \textbf{user}: 6. Jean Galfione (France) 5. 65}\\
    \emph{ \textbf{assistant}: \ul{Name: Jean$|$Galfione\textbackslash n Locations: France\textbackslash n}}

\textbf{JSON Few-Shot Examples}: \\
    \emph{ \textbf{user}:The government of Chad has closed N'Djamena University after two days of protests over grant arrears in which Education Minister Nagoum Yamassoum was held hostage for four hours, state radio said on Friday.}\\
    \emph{ \textbf{assistant}: \ul{\{``name": [``Nagoum Yamassoum"], ``location": [``Chad"]\}}}\\
    \emph{ \textbf{user}: 6. Jean Galfione (France) 5. 65}\\
    \emph{ \textbf{assistant}:\ul{ \{``name": [``Jean Galfione"], ``location": [``France"]\}}}

\subsection{Postprocess}
Here is the core code of the postprocessing function for two prompt styles.

\textbf{Serial Style}

\begin{lstlisting}
import re
# Output sample from Llama 2 13B Chat on CoNLL2003 test split
output_content = " Names: name1\nLocations: Villeurbanne"
def postprocess(t: str)
    name = re.findall(r"Names: (.*?)\n", content)
    location = re.findall(r"Locations: (.*?)\n", content)
    if not location: 
        # output may end without \n
        location = re.findall(r"Locations: (.*?)", content)
    return {'name': name, 'location': location}
postprocess(output_content)
\end{lstlisting} 

\textbf{JSON Style}

\begin{lstlisting}
import json
# Output sample from Llama 2 13B Chat on CoNLL2003 test split
output_content = " Sure, I can do that! Here's the output based on the text you provided:\n\nOutput: {\"name\": [], \"location\": []}"
def postprocess(t: str)
    while t[0] != "{": t = t[1:]
    while t[-1] != "}":  t = t[:-1]
    predict = json.loads(t)
    return predict
postprocess(output_content)
\end{lstlisting}

\section{Supplementary Material: Political Ideology Prediction}

\subsection{Dataset}

The specific keywords for each dataset for Twitter API are listed below:

\noindent \textbf{2020 US Election} {\scriptsize `JoeBiden', `DonaldTrump', `Biden', `Trump', `vote', `election', `2020Elections', `Elections2020', `PresidentElectJoe', `MAGA', `BidenHaris2020', `Election2020'}

\noindent \textbf{CAD COVID-19} {\scriptsize `trudeau', `legault', `doug ford', `pallister', `horgan', `scott moe', `jason kenney', `dwight ball', `blaine higgs', `stephan mcneil', `cdnpoli', `canpol', `cdnmedia', `mcga', `covidcanada' and all combinations of `covid' or `coronavirus' as the prefix and the (full \& abbreviated) name of each provinces and territories as the suffix}

\noindent \textbf{CAD COVID-19} {\scriptsize `trudeau', `legault', `doug ford', `pallister', `horgan', `scott moe', `jason kenney', `dwight ball', `blaine higgs', `stephan mcneil', `cdnpoli', `canpol', `cdnmedia', `mcga', `covidcanada' and all combinations of `covid' or `coronavirus' as the prefix and the (full \& abbreviated) name of each provinces and territories as the suffix}

\noindent \textbf{2021 CAD Election} {\scriptsize `trudeau', `freeland', `o’toole', `bernier', `blanchet', `jagmeet singh', `annamie', `debate commission', `reconciliation', `elxn44', `cdnvotes', `canvotes', `canelection', `cdnelection', `cdnpoli', `canadianpolitics', `canada', `forwardforeveryone', `readyforbetter', `securethefuture', `NDP2021', `votendp', `orangewave2021', `teamjagmeet', `UpRiSingh', `singhupswing', `singhsurge', `VotePPC', `PPC', `peoplesparty', `bernierorbust', `mcga', `saveCanada', `takebackcanada', `maxwillspeak', `LetMaxSpeak', `FirstDebate', `frenchdebate', `GovernmentJournalists', `JustinJournos', `everychildmatters', `votesplitting', `ruralcanada', `debatdeschefs', `électioncanadienne', `polican', `bloc', `jevotebloc'}

\subsection{Politically Explicit Keywords}

For the US, we have two parties, Democrat and Republican.

\noindent \textbf{Democrat} {\scriptsize `liberal,' `progressive,' `democrat,' `biden'}

\noindent \textbf{Republican} {\scriptsize `conservative,' `gop,' `republican,' `trump,'} \\

For the CAD,  we have five parties, CPC, GPC, LPC NDP and PPC.

\noindent \textbf{CPC} {\scriptsize `erin o'toole', `andrew scheer', `conservative', `conservative party', `cpc', `cpc2021', `cpc2019', `conservative party of canada'}

\noindent \textbf{GPC} {\scriptsize `annamie paul', `green party', `gpc', `gpc2019', `gpc2021', `green party of canada'} 

\noindent \textbf{LPC} {\scriptsize `justin trudeau', `liberal', `liberal party', `lpc', `lpc2021', `lpc2021', `lpc2019', `liberal party of canada'}

\noindent \textbf{NDP} {\scriptsize `jagmeeet singh', `new democrat', `new democrats', `new democratic party', `ndp', `ndp2021', `ndp2019'}

\noindent \textbf{PPC} {\scriptsize `maxime bernier', `people's party', `ppc', `ppc2019', `ppc2021', `people's party of canada'}

\subsection{Implicit Political Ideology Prediction}
For the implicit political ideology prediction task, we use the predicted users from the trained explicit political ideology task on the profiles as the (weak) labels for the explicit political ideology prediction task. Table \ref{table:US_UPA}, \ref{table:CAD_COVID-19_UPA}, and \ref{table:2021_CAD_Election_UPA} show the number of weak labels we can extract.We fine-tune a RoBERTa-base model on each dataset's tweets on the masked language modeling task. We then embed each tweet using the pre-trained RoBERTa-base model. Each tweet is represented by a 768-dimensional vector. Since these tweets are political, but not politically explicit, we filter users with a minimum number of tweets. We determine the filter of users amongst 1, 3, 5, 10, 15, 20 and 25 using 5-fold CV to be 10 for the 2020 US
Election dataset and 5 for both the CAD COVID-19 and the 2021 Canadian Election dataset. 

\begin{table}[ht!]
\centering
\begin{tabular}{l|rrr}
Party & Support & F1-Score & \# of Users \\
 \hline
Republican & 854 & 97.21 ± 0.66 & 86,989 \\
Democrat   & 928 & 97.40 ± 0.63 & 82,923 \\
\end{tabular}
\caption{US 2020 Election Party Affiliation Classification. Cohen Kappa score of 0.76. Total 763,164 users.}
\label{table:User_Party_Affiliation_Classification_Accuracy}
\end{table}

\begin{table}[ht!]
\centering
\begin{tabular}{l|rrr}
Party & Support & F1-Score & \# of Users \\
 \hline
Republican & 854 & 97.21 ± 0.66 & 86,989 \\
Democrat   & 928 & 97.40 ± 0.63 & 82,923 \\
\end{tabular}
\caption{2020 US Election Party Affiliation Classification. Cohen Kappa score of 0.76.}
\label{table:US_UPA}
\end{table}

\begin{table}[ht!]
\centering
\begin{tabular}{l|rrr}
Party & Support & F1-Score & \# of Users \\
\hline
CPC &  98 & 92.93 ± 1.12 &  1,769 \\
GPC &  60 & 88.50 ± 1.54 &  97 \\
LPC &  100 & 90.89 ± 1.34 & 783 \\
NDP &  124 & 93.44 ± 0.53 & 370 \\
NO\_PARTY & 105 & 86.16 ± 1.97 & 667 \\
PPC & 95 & 94.09 ± 1.34 & 402 \\
\end{tabular}
\caption{CAD COVID-19 Party Affiliation Classification. Cohen Kappa score of 0.74}
\label{table:CAD_COVID-19_UPA}
\end{table}

\begin{table}[ht!]
\centering
\begin{tabular}{l|rrr}
Party & Support & F1-Score & \# of Users \\
\hline
CPC & 183 & 97.82 ± 0.39 &  6437 \\
GPC & 57 & 97.22 ± 1.11 &  7 \\
LPC & 152 & 94.60 ± 0.74 & 5117 \\
NDP & 71 & 98.32 ± 1.07 & 108 \\
NO\_PARTY & 28 & 66.64 ± 6.10 & 629 \\
PPC & 67 & 96.74 ± 1.68 & 44 \\
\end{tabular}
\caption{2021 CAD Election Party Affiliation Classification. Cohen Kappa score of 0.61}
\label{table:2021_CAD_Election_UPA}
\end{table}

\subsection{Prompts}

For our political ideology prediction, we use the following prompting style. In the system prompt, we replace the \textit{country} with either ``US'' or ``Canada'' depending on the dataset. For the ``US'' dataset, the \textit{parties} are replaced with [``Democrat'', ``Republican''.] For the ``Canadian'' datasets, the \textit{parties} are replaced with [``CPC'', ``GPC'', ``LPC'', ``NDP'', ``PPC'']. The \textit{data\_type} can be either ``profile'' or ``tweets''.

\textbf{System}: \emph{You are a political scientist for \{\{country\}\}. You must classify the user as one of the following parties: \{\{parties\}\} based on provided \{\{data\_type\}\}. Do not include any other text or explanation.}

For all our prompts, the user prompt would be the following. Like before, \textit{data\_type} can be either ``profile'' or ``tweets''. For the ``profile'', the ``data'' would be the Twitter user profile. For the ``tweets'', the ``data'' would be the Twitter tweets from the user. We provide the tweets in the form of a numbered list, with a new line in-between.

\textbf{User}: \emph{\{\{data\_type\}\}: \{\{data\}\}}.

After the system prompt, if the prompting is for few-shot, we inject two examples per class.  

\textbf{User}: \emph{\{\{data\_type\}\}: \{\{example data\}\}}.

\textbf{Assistant}: \emph{\{\{example data answer\}\}}.

\begin{table*}[ht!]
\centering
\begin{tabular}{ll|lll}
Model & Method & US Election & CAD COVID-19 & CAD 2021 Election\\ \hline
Llama 2 (13B) & \multirow{4}{*}{Zero-Shot} & 86.6 ± 0.5 & 88.6 ± 1.1 & 80.8 ± 1.6 \\
Llama 2 (70B) &  & 95.0 ± 0.5 & 89.0 ± 1.1 & 80.4 ± 1.6 \\
GPT 3.5 & & 97.6 ± 0.6 & 92.4 ± 0.9 & 83.7 ± 1.2 \\
GPT 4.0 & & \textbf{98.1 ± 0.4} & 94.8 ± 0.7 & 86.1 ± 1.4 \\
\hline
Llama 2 (13B) & \multirow{4}{*}{Few-Shot} & 95.5 ± 1.1 & 90.2 ± 0.9 & 82.1 ± 1.6 \\
Llama 2 (70B) &  & 96.3 ± 0.5 & 92.5 ± 1.3 & 85.2 ± 1.0 \\
GPT 3.5 &  & 97.0 ± 0.8 & 94.7 ± 0.8 & 87.7 ± 1.3 \\
GPT 4.0 &  & 97.6 ± 0.5 & \textbf{95.1 ± 0.6} & 89.4 ± 1.2 \\ \hline
RoBERTa & Finetuning & 97.3 ± 0.6 & 91.2 ± 0.2 & \textbf{95.2 ± 0.7}
\end{tabular}
\caption{Text Classification: Explicit Political Ideology Prediction on Profiles}
\label{tab:Profile_Prediction_Results}
\end{table*}

\begin{table*}[ht!]
\centering
\begin{tabular}{ll|lll}
Model & Method & US 2020 Election & CAD COVID-19 & CAD 2021 Election \\ \hline
Llama 2 Chat (13B) & \multirow{3}{*}{Zero-Shot} & 71.9 ± 1.9 & 44.6 ± 1.6 & 48.8 ± 3.5 \\
Llama 2 Chat (70 B) & & 77.2 ± 1.0 & 53.9 ± 1.5 & 55.7 ± 3.3 \\
GPT 3.5 &  & 92.0 ± 0.6 & 59.7 ± 2.0 & 65.0 ± 1.7 \\ \hline
Llama 2 (13B) & \multirow{3}{*}{Few-Shot} & 63.7 ± 1.2 & 32.1 ± 1.9 & 26.6 ± 1.6 \\
Llama 2 (70 B) &  & 44.2 ± 1.9 & 22.5 ± 3.3 & 30.8 ± 2.5 \\
GPT 3.5 &  & 92.9 ± 0.5 & 65.9 ± 2.0 & 75.4 ± 1.6 \\ \hline
RoBERTa + MLP & Fine-Tune & \textbf{93.0 ± 0.2} & \textbf{70.0 ± 2.7} & \textbf{82.3 ± 1.1}
\end{tabular}
\caption{Text Classification: Implicit Political Ideology Prediction on Tweets}
\label{tab:Tweet_Prediction_Results}
\end{table*}

\section{Supplementary Material: Misinformation Detection}

\subsection{Prompt}

For all the misinformation experiments, we use the following prompt. We do not use a system prompt in this case. We replace \textit{STATEMENT} for each instance in the respective dataset:

\textbf{User}: \emph{Rate the truthfulness of the following statement: \{\{STATEMENT\}\}\ Provide a score from 0 to 100, where 0 represents definitively false and 100 represents definitively true. Do not provide any explanations, only respond with the numerical score.'}.

\subsection{Additional Results}

In Table~\ref{tab:misinfo_results}, we examine the upper potential of this score-based approach. Zero-shot results results are reprinted from the main paper table. The remainder are each the result of one evaluation run on the full test set. In particular, the Oracle results show the performance with the ideal score threshold for converting to categorical labels (one threshold for LIAR, two thresholds for CT-FAN-22). This optimization is not possible in the real world, but it shows the maximum achievable result with this approach. Then, we report the result of tuning the threshold on the validation set of LIAR for GPT 4 (Val Tuned), which unlike Oracle might be done in a practical, non-transfer setting. Finally, we compare the results of RoBERTa with fine-tuning vs. transfer on CT-FAN-22.

The oracle results show that although GPT 3.5 does better zero-shot, GPT 4 has higher potential. Furthermore, this potential might be realizable if tuning on a validation set is possible - Val Tuned GPT-4 nearly matches Oracle GPT-4 on LIAR.

\begin{table*}[ht!]
\centering
\begin{tabular}{ll|ll}
Model & Method & LIAR & CT-FAN-22 \\ \hline
Llama 2 Chat (13B) & \multirow{4}{*}{Zero-Shot} & 50.0 ± 1.3 &  21.2 ± 3.2 \\
Llama 2 Chat (70 B) &  & 49.1 ± 2.5 & 25.4 ± 2.1  \\
GPT 3.5 &  & 66.3 ± 2.1 & 43.7 ± 1.9  \\ 
GPT 4 &  & 61.5 ± 2.1 & 42.0 ± 2.6 \\ \hline
Llama 2 Chat (13B) & \multirow{4}{*}{Oracle Tuned} & 56.4 & 28.8  \\
Llama 2 Chat (70 B) &  & 56.4 & 28.0  \\
GPT 3.5 &  & 67.5 &  43.9 \\ 
GPT 4 &  & 68.7 &  50.3\\ \hline
GPT 4 & Val Tuned & 68.2 & -- \\ \hline
RoBERTa & Fine-Tune & 64.7 & 23.0 \\
RoBERTa & Transfer & -- & 26.8 
\end{tabular}
\caption{Text Classification: Misinformation. Even in idealistic conditions, we see that both Llama and RoBERTa are lacking in this challenging task.}
\label{tab:misinfo_results}
\end{table*}

\end{document}